\def\year{2021}\relax
\documentclass[letterpaper]{article} 
\usepackage{aaai21}  
\usepackage{times}  
\usepackage{helvet} 
\usepackage{courier}  
\usepackage[hyphens]{url}  
\usepackage{graphicx} 
\usepackage{natbib}  
\usepackage{caption} 
\usepackage{subfigure}
\usepackage{enumitem}
\usepackage{comment}

\usepackage{xcolor}

\usepackage{soul}
\usepackage{url}
\usepackage{hyperref}
\usepackage[utf8]{inputenc}
\usepackage{caption}
\usepackage{graphicx}
\usepackage{amsmath}
\usepackage{amsfonts}
\usepackage{booktabs}
\usepackage{tabu}
\usepackage{fancyhdr}
\usepackage{algorithm}
\usepackage{setspace}
\usepackage{algcompatible}
\usepackage[noend]{algpseudocode}
\makeatletter
\def\BState{\State\hskip-\ALG@thistlm}
\makeatother

\algdef{SE}[SUBALG]{Indent}{EndIndent}{}{\algorithmicend\ }%
\algtext*{Indent}
\algtext*{EndIndent}

\fancypagestyle{firstpage}{%
}

\title{Inferring COVID-19 Biological Pathways from Clinical Phenotypes via Topological Analysis}

\author{
    Negin Karisani\textsuperscript{\rm 1}, 
    Daniel E. Platt\textsuperscript{\rm 2},
    Saugata Basu\textsuperscript{\rm 1}\thanks{S. Basu and N. Karisani were partially supported by NSF grant DMS-1620271.},
    Laxmi Parida\textsuperscript{\rm 2}
}
\affiliations{

    \textsuperscript{\rm 1} Purdue University \\
    \textsuperscript{\rm 2} IBM T.J. Watson Research Center\\
    parida@us.ibm.com
}

\begin{document}

\maketitle

\begin{abstract}

COVID-19 has caused thousands of deaths around the world and also resulted in a large international economic disruption. Identifying the pathways associated with this illness can help medical researchers to better understand the properties of the condition. This process can be carried out by analyzing the medical records. It is crucial to develop tools and models that can aid researchers with this process in a timely manner. However, medical records are often unstructured clinical notes, and this poses significant challenges to developing the automated systems. In this article, we propose a pipeline to aid practitioners in analyzing clinical notes and revealing the pathways associated with this disease. Our pipeline relies on topological properties and consists of three steps: 1) pre-processing the clinical notes to extract the salient concepts, 2) constructing a feature space of the patients to characterize the extracted concepts, and finally, 3) leveraging the topological properties to distill the available knowledge and visualize the result. Our experiments on a publicly available dataset of COVID-19 clinical notes testify that our pipeline can indeed extract meaningful pathways. 


\end{abstract}

\section{Introduction} \label{sec:intro}
Since the early stages of the COVID-19 pandemic, the scientific community has made tremendous effort to address the clinical course of the virus.
However, there is still a lot to reveal about COVID-19. For instance, most people who contract COVID-19 develop mild to moderate symptoms \cite{who}, some may show no symptoms, while for others the disease can be fatal. To better understand different strains of COVID-19 one approach is to study the underlying pathways. The aim of this study is to investigate the application of topological properties in automatically inferring candidate pathways. We use unstructured clinical notes as the source of information to automatically extract phenotypes to be used in our topological model. Phenotypes are the symptoms and signs that reflect the presence of a disease--in the follows, we refer to them as symptoms.

Advancement in technology has helped scientists to garner large amount of biomedical data. This has provided the community with unprecedented opportunities to study and better understand the spread of diseases. However, this burst of information has posed significant challenges to the traditional data analysis and visualization techniques. Traditional infographics, such as Venn diagrams which are still widely used to compare and contrast set of symptoms, fail to aid practitioners in analyzing large set of symptoms. Thus, tools that can effectively employ the techniques in other scientific communities to facilitate this process are of great value.


In this article, we rely on concepts from Topological Data Analysis and propose a pipeline to automatically extract candidate pathways associated with COVID-19 from clinical notes. Our pipeline which is based on the notion of Redescriptions \cite{Parida2005RedescriptionMS,MULLINS2006,Platt2016} consists of three steps: 1) pre-processing the notes and identifying the candidate symptoms, 2) mapping the symptoms to the space of the patients, and finally, 3) extracting the topological properties and their visualization. We have evaluated our pipeline in a publicly available dataset of COVID-19 clinical notes. The results show that our model is able to extract meaningful pathways. For example, in Section \ref{sub-sec:result} we demonstrate that there are potentially distinctive pathways between coughers and non-coughers.

The remainder of this article is organized as follows: in Section~\ref{sec:background} we provide an overview of the concepts that we use. In Section~\ref{sec:pipeline} we present our pipeline and in Section~\ref{sec:implementation} we discuss the implementation detail. In Section~\ref{sec:experimental} we explain the detail of our experiments and in Section~\ref{sec:results} we present and discuss the results. Finally, in Section~\ref{sec:conclusion} we conclude the article.  


\section{Background} \label{sec:background}
In this section, we review the concepts used in the remainder of the paper.

\subsection{Redescriptions}
\label{sec:background-redes}
Redescriptions are used to identify the phenomena that occur in different ways. The concept was first introduced in \cite{10.1145/1014052.1014083}, and later in \cite{Parida2005RedescriptionMS} was generalized to a framework called redescription mining, for which the authors present some applications in Genome Ontology database.

Redescriptions are mathematically formalized using Boolean algebra, which is also used to model the cause-effect relationship among the symptoms. Two different sets of symptoms which correspond to the same group of patients is an example of redescriptions. More specifically, suppose $s_1$, $s_2$ are two symptoms, and $P_1$, $P_2$ their respective set of patients. If the presence of symptom $s_1$ implies the presence of symptom $s_2$, then $P_1 \subseteq P_2$. If we consider the combination of the symptoms (i.e. $s_1 \wedge s_2$), then the group of the patients who experience both symptoms is $P_1 \cap P_2 = P_1$, which is the same group of patients that we obtain by considering only the symptom $s_1$. Redescriptions--the combination of symptoms that give rise to the same group of patients--can reveal logical associations among symptoms. They can highlight the underlying pathways and are commonly used to derive rules in the pathways \cite{MULLINS2006}.

\subsection{Topological Data Analysis}
Over the past two decades, Topological Data Analysis (TDA), raised from Algebraic Topology, has found its way in the real-world applications.
In \cite{Dagliati-2019}, TDA is used to model disease progression by inferring temporal phenotypes.
More recently \cite{Wang-2020} use TDA and machine learning techniques to investigate genome mutation of SARS-COV-2.
Some other examples in biology include analysis of
brain neural activities \cite{dabaghian_topological_2012,8999328}, and cancer genomics \cite{nicolau_topology_2011,Rabadan2020}. In this section, we aim to provide a brief overview of the primary TDA concept, i.e., the persistent homology. We avoid the mathematical detail which is beyond the scope of this article. For a thorough description see \cite{Edelsbrunner_survey,wasserman-2018}.

Let $M$ be a continuous space equipped with a metric $\delta$, the topological invariants of $M$ are defined as the properties that do not change under continuous deformation (i.e. twisting but not tearing). The invariants in $M$ for lower dimensions are usually referred to as the connected components, the holes, and the void spaces, respectively in dimension 0, 1 and 2; in the higher dimensions, they are understood as $k$-dimensional holes. The number of $k$-dimensional holes in $M$ are called the $k$-th betti numbers.

Given data points $X$ and a distance function $\delta$---$X$ represents as the points sampled from $M$---the goal is to compute the topological invariants of the underlying structure of $X$ (i.e. space $M$). A common approach to accomplish this is by constructing $k$-simplexes over $X$. Intuitively, one could think of a $k$-simplex as the smallest convex hull of $k+1$ points. A collection of $k$-simplexes glued together is called a simplicial complex (satisfying some conditions). Since it is not reasonable to begin with all the possible $k$-simplexes over the data points in $X$; the technique is to add the simplexes in a sequence of steps. 
First, a parameter is selected and  the initial simplicial complex $S$, is set to be the collection of points in $X$ as the 0-simplexes; then the parameter is increased such that at each step just a specific set of simplexes, that satisfy some conditions, could be added to $S$; this procedure creates a filtration of simplicial complexes on $X$, which then is analyzed. The conditions to select a subset of simplexes at each step, give rise to different types of simplicial complexes.  

\begin{figure}[H]
	\centering
	\subfigure[]{%
		\label{Fig:1}%
		\includegraphics[height=0.54in]{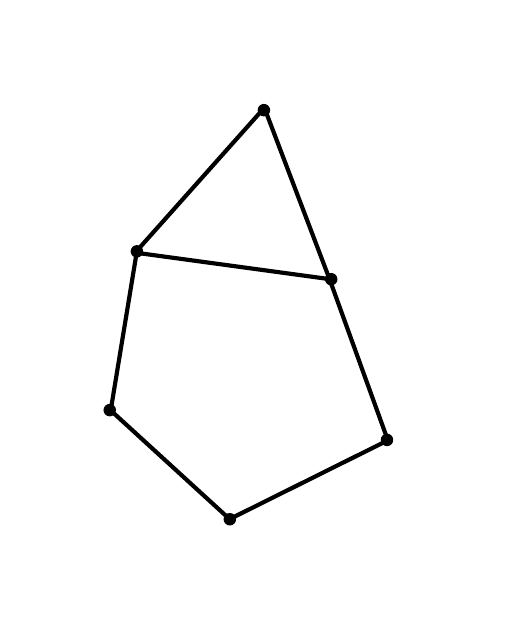}}%
	\qquad
	\subfigure[]{%
		\label{Fig:2}%
		\includegraphics[height=0.54in]{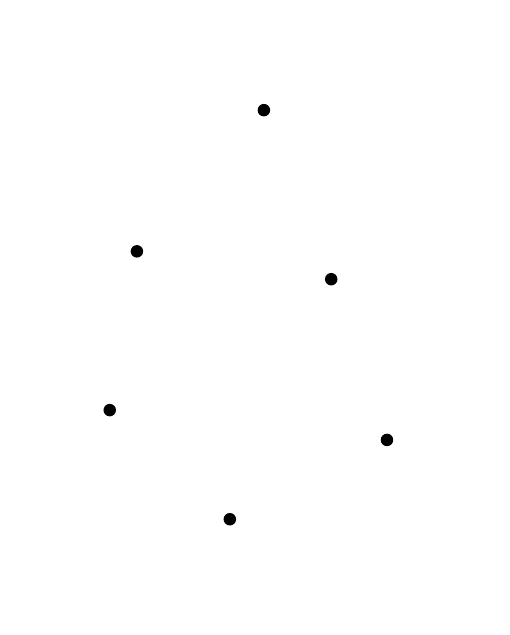}}%
	\qquad
	\subfigure[]{%
		\label{Fig:3}%
		\includegraphics[height=0.54in]{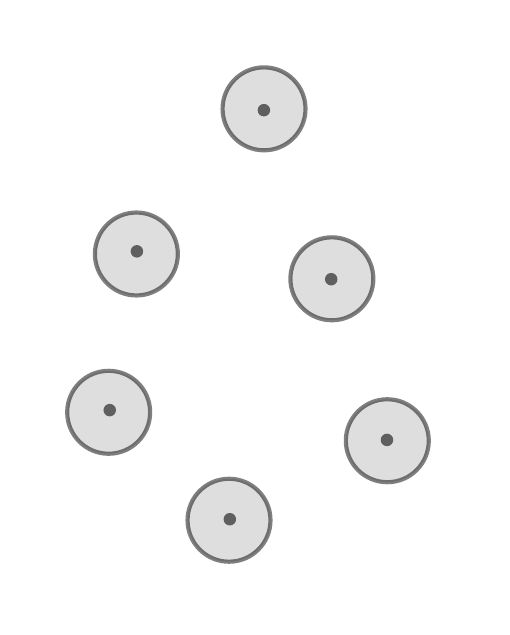}}%
	\qquad
	\subfigure[]{%
		\label{Fig:4}%
		\includegraphics[height=0.54in]{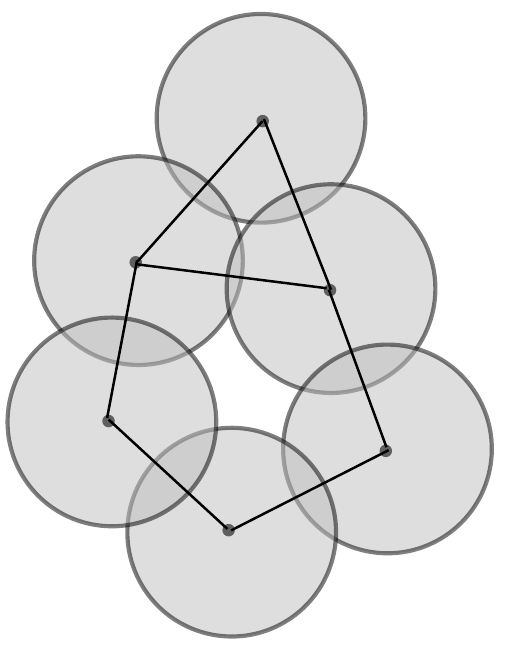}}%
	\qquad
	\subfigure[]{%
		\label{Fig:5}%
		\includegraphics[height=0.54in]{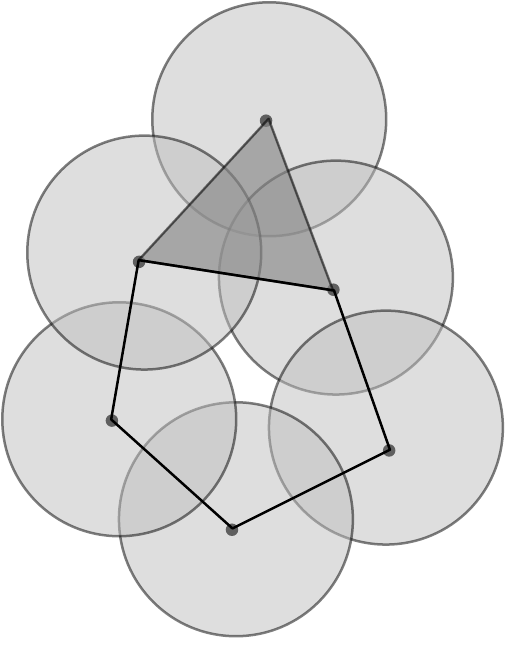}}%
	\caption{\small Recovering topological properties using simplicial complexes}\label{Fig:psh}
	\vspace*{-.2cm}
\end{figure} 

An example of a simplicial complex is \v{C}ech complex.  
Figure~\ref{Fig:psh} shows a simple illustrative example. The goal is to recover the topological invariants of the space in Figure~\ref{Fig:1}. It is clear that the 0-betti number is one, since there is only one connected component; and the 1-betti number is two, since there are two holes; the higher dimensional betti numbers are zero. The dataset $X$ is given by the six sampled points in Figure~\ref{Fig:2}. To construct the \v{C}ech complex over $X$, we begin with the points in $X$ as 0-simplexes. In order to construct the higher dimensional simplexes, we start growing a ball at each point, as in Figure~\ref{Fig:3}; at this state, the 0-betti number is six, and 1-betti number is zero. By increasing the radius, some of the balls start to overlap each others; for each $k+1$ overlapped balls we insert a $k$-simplex. Figure~\ref{Fig:4} shows a collection of 1-simplexes, line segments joining the two points, created by the pairwise overlap of their corresponding balls; what can be clearly seen in this simplicial complex is that it recovers the topological properties of the underlying structure in Figure~\ref{Fig:1}. If we increase the radius further, the three balls at the top begin to overlap each others, hence we can add a 2-simplex---a filled in triangle---as in Figure~\ref{Fig:5}. Therefore, the hole which was created at Figure~\ref{Fig:4} disappeared by increasing the radius at Figure~\ref{Fig:5}, as a result that topological property is lost; this could eventually happen for the second hole as we continue to increase the radius and add more simplexes. It is important to notice that the topological properties that persist for a longer period, before they disappear, best represent the properties of the underlying structure. This characteristic is the basic principle of the persistent homology method, which was first formalized in \cite{Edelsbrunner2002}.

A diagram known as Barcode is commonly used to keep track of the lifetime of topological properties. In the barcode, each topological property is represented as a horizontal line segment. The line segments span the period that the corresponding topological properties exist, along the parameter axis (i.e. radius). We use the barcode in Section~\ref{sub-sec:result} (see Figure~\ref{Fig:barcode}).  

\section{Proposed Pipeline} \label{sec:pipeline}
In this section, we introduce our pipeline. The first step is to extract structured data---the set of symptoms and their corresponding patients---from the unstructured clinical records. Next step is to define the feature space, the sampling strategy, and the metric to measure the similarity between the data points. Finally, the topological properties are extracted and are visualized.

\noindent\textbf{Concept extraction}: We carry out the concept extraction in three steps: 1) We parse the clinical notes and map the biological terms to the concepts in a medical ontology. 2) Since the clinical notes have informal language model their parsing can be noisy. Thus we ask the user to cure the candidate relations and resolve the inconsistencies. 3) We use the health records to construct an association matrix between the patients and the extracted concepts. 

Natural language processing techniques are widely used to analyze biomedical documents \cite{DEMNERFUSHMAN2009760,10.1093/jamia/ocy173}. Despite the significant advances in neural text processing over the last decade, we found that the existing methods are not adequate to effectively parse the medical records. Thus, to reduce the noise and ensure that the extracted terms are indeed valid medical concepts we use manual supervision to validate the automatic process.

\noindent\textbf{Feature space construction}:
In our model, features correspond to the patients, and the data points correspond to the combination of concepts--we call them patterns. Given a feature vector--i.e., a data point--a feature is set to 1 if the corresponding patient shows all the symptoms associated with the data point. Thus, a data point is understood as a cluster of patients, who share the same set of symptoms--i.e., pattern. 
 
 As mentioned in Section~\ref{sec:background-redes}, pathways can be inferred by identifying the redescriptions---i.e., patterns that have the same group of patients. However, in order to make inference about the underlying pathways, it is important to analyze the patterns whose clusters are statistically significant.

 Since tests such as Binomial are not successful in separating higher-order correlations, which can distinguish groups of patients that identify disease processes from the impact of pair-wise correlations, we use cumulant correlation expansions. In quantum field theory, they emerge as connected Feynman diagrams (1-particle irreducible). These are multivariate moments related to cumulants appearing in statistics. In that context, their generating functions factor according to partitions of sets.

Let $G_\bullet$ represent the moments 
in moment generating functions $\mathbb{E}\left[\exp \left( \sum_i x_i J_i\right) \right]$ where the $J_i$'s are the conjugate variables, and $\Gamma_\bullet$ represent the higher dimensional cumulants, e.g. for the symptoms $x_i, x_j, x_k$ then $G_{ij} = E(x_i x_j)$ and $G_{ijkk} = E(x_i x_j x_k^2)$, and $\Gamma_{ij}$ and $\Gamma_{ijkk}$ are the corresponding cummulants. 
The factorizations are as follows. 
{\small
	\begin{equation*}
		\begin{split}
			\mathbb{E}\left[\exp \left(\sum_i x_i J_i\right)\right] 
			= A + \sum_i J_i G_i + \frac{1}{2!}\sum_{ii'} J_i J_{i'} G_{ii'} + \\
			\frac{1}{3!}\sum_{ii'i''} J_i J_{i'} J_{i''} G_{ii'i''} +\\	
			\frac{1}{4!}\sum_{ii'i''i'''} J_i J_{i'} J_{i''} J_{i'''} G_{ii'i''i'''} + \cdots  \\	 
			= \exp\left(
			\sum_i J_i \Gamma_i +\frac{1}{2!}\sum_{ii'} J_i J_{i'} \Gamma_{ii'} + \frac{1}{3!}\sum_{ii'i''} J_i J_{i'} J_{i''} \Gamma_{ii'i''} + \right. \\	
			\left. \frac{1}{4!}\sum_{ii'i''i'''} J_i J_{i'} J_{i''} J_{i'''} \Gamma_{ii'i''i'''}
			+ \cdots\right),
		\end{split}
	\end{equation*}
}
where $A$ is nominally 1, seen by setting the $J_i = 0$.

The power series in the $J_i$'s then require
{\small  
	\begin{align*}
		G_k &= \Gamma_k \nonumber \\
		G_{kk'} &= \Gamma_{kk'} + \Gamma_k \Gamma_{k'} \nonumber \\
		G_{kk'k''} 
		& = 
		\Gamma_{kk'k''} 
		+ \Gamma_k \Gamma_{k'k''} + \Gamma_{k'}\Gamma_{k''k}  +	\Gamma_{k''}\Gamma_{kk'}
		+ \Gamma_k \Gamma_{k'} \Gamma_{k''} \\ 
		G_{kk'k''k'''}  &= 
		\Gamma_{kk'k''k'''} 
		+ 
		\Gamma_k \Gamma_{k'k''k'''} + \Gamma_{k'}\Gamma_{k''k'''k} + \nonumber \\
		&  
		\Gamma_{k''}\Gamma_{k'''kk'} + \Gamma_{k'''}\Gamma_{kk'k''}
		+ 
		\Gamma_{k'''k'}\Gamma_{k''k} + \\
		& \Gamma_{k'k''}\Gamma_{k'''k} + \Gamma_{k'''k''}\Gamma_{kk'} + 
		2 \Gamma_k \Gamma_{k'} \Gamma_{k''k'''} + \\ & 2\Gamma_k \Gamma_{k'''} \Gamma_{k'k''} 
		+ 2 \Gamma_k \Gamma_{k''} \Gamma_{k'k'''} + 2\Gamma_{k'}\Gamma_{k'''}\Gamma_{kk''} + \nonumber \\
		&
		2 \Gamma_{k'}\Gamma_{k''}\Gamma_{kk'''} + 2\Gamma_{k'''}\Gamma_{k''}\Gamma_{kk'}
		+\Gamma_k\Gamma_{k'}\Gamma_{k''}\Gamma_{k'''} \nonumber
	\end{align*}\label{eqn:gamma}
}

 	We apply the above factorization to the clusters, and shuffle the symptoms to test significance by constructing variances and null hypotheses.
 
   To search for the redescriptions, we need to investigate the cause--effect relationships among the selected patterns.
   However, often due to the misclassifications of patients, e.g., caused by wrong diagnosis, the set inclusion property does not hold in the data. Therefore, the exact equality of sets should be estimated. This estimation can be done by Jaccard distance, which measures the dissimilarity between sets. For the two sets $A$ and $B$, Jaccard distance is defined by,   

$$d(A, B) = 1 - \frac{\vert A \cap B\vert}{\vert A \cup B\vert}.$$

For the example in Section~\ref{sec:background-redes}, when $P_1 \subseteq P_2$, then the Jaccard distance $d(P_1 \cap P_2, P_1) = 0$, otherwise if $P_1 \not\subseteq P_2$ then $0 < d(P_1 \cap P_2, P_1) \leq 1$, which can be interpreted as the probability that subjects picked from the two sets are not shared. 

Hence, we consider Jaccard distance to measure the distances between the sampled data points.

    \noindent\textbf{Topological analysis and visualization}:
    To explore the structure of the space created in the previous step, Vietoris–Rips (VR) complexes are employed to construct the filtration. The VR complex is an abstract simplicial complex with 0-simplexes as the data points, and $k$-simplexes are created for any $k+1$ points whose pairwise distances are at most $2r$, while $r$ is fixed.
    
    The initial simplicial complex is a collection of 0-simplexes which correspond to the sampled data points--i.e., the clusters of patients selected from the previous step--and Jaccard distance is used as the filtration parameter to construct the VR complexes.
    Finally, the barcode is generated and representative cycles of the bars are retrieved for further analysis. 

\section{Implementation Details} \label{sec:implementation}

To parse the clinical notes and extract the biomedical terms we used Amazon Comprehend Medical (ACM)\footnote{\url{https://aws.amazon.com/comprehend/medical/}}, an online proprietary NLP programming interface to analyze the unstructured clinical notes. For technical details regarding ACM see \cite{ML4H,Bhatia2020-1}. We also used the International Classification of Diseases (ICD-10CM)\footnote{\url{https://www.cdc.gov/nchs/icd/icd10cm.htm}} to select the concepts, which are mapped by ACM to the extracted terms. ICD is a medical ontology, published by the World Health Organization to classify diseases, symptoms, and other medical conditions. 

In the TDA step, we used Dionysus\footnote{https://mrzv.org/software/dionysus2/} package for the construction of simplicial complexes and visualization. We also incorporated the Cyclonysus\footnote{https://github.com/sauln/cyclonysus} implementation to retrieve the representative cycles of the 1-dimensional topological properties. 

\section{Experimental Details} \label{sec:experimental}
We begin this section by describing the dataset, then we discuss the steps of the experiment. 

\subsection{Dataset}
\label{sub-sec:dataset}
We used the dataset introduced in \cite{Xu2020}\footnote{ \url{https://github.com/beoutbreakprepared/nCoV2019/tree/master/latest_data}}. The dataset is continually updated with the available records of confirmed COVID-19 patients. We used the version published on June 8, 2020. Among the available records in the data set we retained all the records that their ``symptom'' field was non-empty, this amounted to 1,545 patients. This field, which is a textual feature, is a clinical note describing the patient's medical state.

\subsection{Experimental Setup}
\label{sub-sec:Details}
ACM associates a list of ICD-10CM codes to each extracted medical condition, ordered by their confidence scores, hence we retained a code with the highest confidence score. We only considered medical conditions that at least 0.3 percent of patients experienced. If the ICD-10CM codes associated to a medical condition were at the same level of the hierarchical ontology and ACM was assigning high confidence scores to all of them, we considered them as one class. An example of that includes $R53. = \{ R53.1: \textit{Weakness},\ R53.81: \textit{Malaise},\ R53.83:\textit{Other fatigue}\}$. We retained the data corresponded to thirty-one ICD-10CM codes. 
Based on the data Fever, Cough and Fatigue are the most common symptoms among the COVID-19 patients. Table~\ref{Table:codes} presents the list of selected classes and their number of patients, and Table~\ref{Table:classsize} provides the number of patients who experienced $k$ medical conditions. 

{	\small
	\begin{table*}
		\begin{center}
			\begin{tabu}{p{1.75in} p{0.75in} p{0.25in} p{0.25in} p{1.75in} p{0.67in}  p{0.25in} p{0.25in} }
				\cline{1-6}
				\hline 
				\centering \textbf{Description} & \textbf{ICD-10CM} & \pmb{$\sharp$}  & \pmb{$\%$} &
				\centering	\textbf{Description} & \textbf{ICD-10CM} & \pmb{$\sharp$} & \pmb{$\%$}
				\\
				\cmidrule(l){1-4} \cmidrule(l){5-8}
				\cmidrule(l){1-4} \cmidrule(l){5-8}
				Acute myocardial infarction & I21.9 &  5 &  0.3 & Chest pain & R07. &  24 &  1.6\\
				Pulmonary heart disease & I27. &  6 &  0.4 & Abnormal sputum & R09.3 &  43&  2.8\\
				Cardiac arrhythmia & I49.9 &  5 &  0.3& Nasal congestion & R09.81 &  11 &  0.7\\
				Heart failure& I50.9 &  9&  0.6& Abdominal pain & R10.9 &  6&  0.4\\
				Acute pharyngitis & J02.9 &  136 &  8.8& Nausea &R11. &  29&  1.9\\
				Pneumonia & J18. &  151&9.7& Diarrhea &R19. &  28&  1.8\\
				Nasal sinuses & J34.89 &  65& 4.2& Dizziness &R42 &  6&  0.4\\
				Respiratory failure & J96. &  64&4.1& Fever &R50.9 &  1073&  69.4\\
				Pain in joint& M25.50 &  23&1.5& Headache &R51 &  76&  5\\
				Muscle spasm & M62.838 &  24&1.6& Unspecified pain&R52 &  24&  1.6\\
				Myalgia & M79.10 &  70& 4.5& Fatigue &R53. &  177&  11.5\\
				Disorders of bone & M89.8X9 &  10& 0.6&Anorexia &R63.0 &  8& 0.5\\
				Kidney failure & N17.9 &  9& 0.6& Sepsis& R65.21 &  17&  1.1\\
				Cough & R05 &  594& 38.4& Chills &R68.83 &  41&  2.7\\
				Abnormalities of breathing  & R06. &  138  &9& Dry mouth &R68.2 &  6&  0.4\\
				Sneezing & R06.7 &  17& 1.1& &&&\\		
				[2pt]\hline
			\end{tabu}
		\end{center}
		\caption{Thirty-one ICD10-CM concepts with the number of patients in each class and their respective percentage of total.} \label{Table:codes}
	\end{table*}

\begin{table}
	\begin{center}
		\begin{tabular}{r@{\quad\quad}l@{\quad}l}
			\hline
			\multicolumn{1}{l}{
				\pmb{$ k$}}&\multicolumn{2}{l}{{\quad}\pmb{$\sharp$}}\\[2pt]
			\hline
			1 & &651 \\ 
			2 &  &431  \\ 
			3 & &286\\ 
			4 & &115\\ 
			5&  &45 \\ 
			6& &11 \\ 
			7& &5  \\ 
			8& & 1 \\
			[2pt]
			\hline
		\end{tabular}
	\end{center}
	\caption{Number of patients with $k$ symptoms.} \label{Table:classsize}
\end{table}
}

In the second step of the pipeline, we selected 632 data points with patterns corresponded to the subsets of the thirty-one ICD-10CM codes. To construct the VR filtration, we set the threshold of the filtration parameter to $0.5$.

\section{Results} \label{sec:results}
In this section, we report the main result and discuss its significance. 

\subsection{Main Result} \label{sub-sec:result}
We obtained topological properties of dimensions 0 and 1; there was no topological property of higher dimensions. We report an important 1-dimensional property which is striking.

As mentioned in Section~\ref{sec:pipeline}, we used Jaccard distance. Therefore, at any two data points, the lower the distance, the more similar their sets of patients are. Following from this, the topological properties whose 1-simplexes corresponded to low distances were of interest.  

\begin{center}
	
	\begin{figure}[H] \centering
		\includegraphics[scale=0.4]{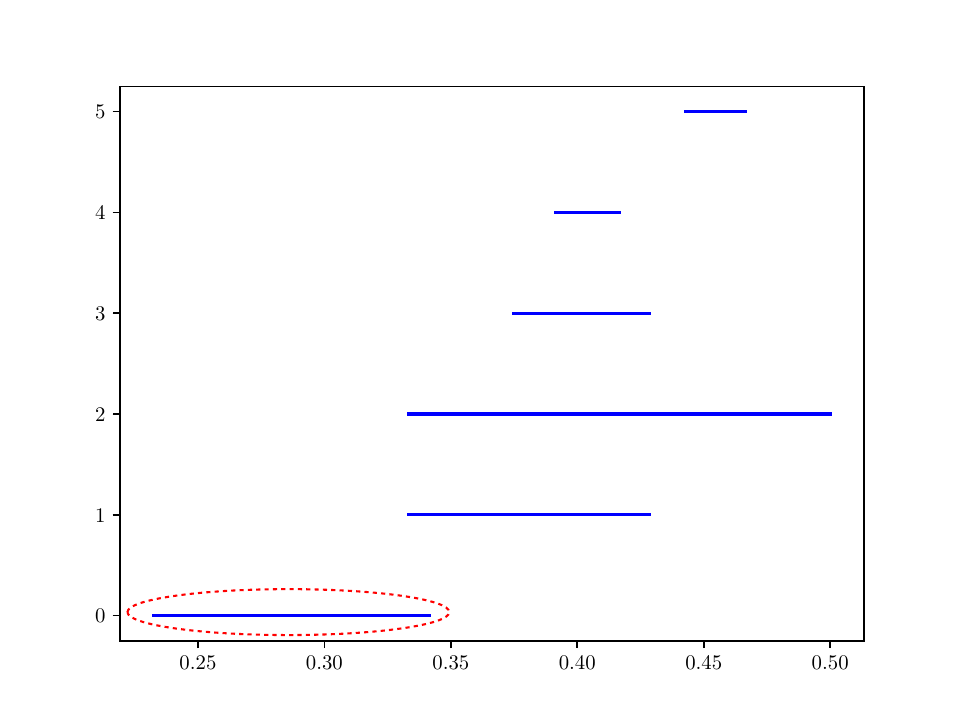} 
		\caption{\small Barcode of 1-dimensional topological properties.}\label{Fig:barcode}
	\end{figure}
\end{center}

Figure~\ref{Fig:barcode} shows the barcode of the 1-dimensional topological properties, whose lifetime is within the interval (0, 0.5). The horizontal axis corresponds to the parameter of the filtration---Jaccard distance---and the vertical axis corresponds to the number of properties. With respect to the previous paragraph, what stands out in the diagram is the first bar annotated by the circled line, which spans between $0.23$ and $0.34$. 

\begin{center}
	\begin{figure}[H] \centering
		\includegraphics[scale=0.6]{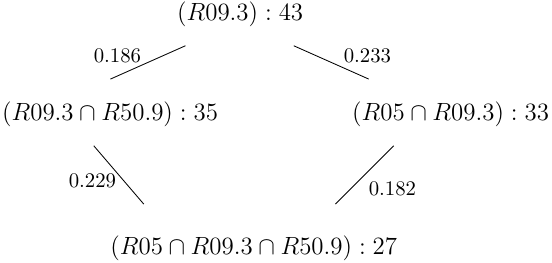}
		\caption{\small Representative cycle of the annotated bar.}\label{Fig:cycle}
	\end{figure}
\end{center}

 Since the 1-dimensional topological properties are understood as the holes made up of points and 1-simplexes, a cycle generating the annotated bar is shown in Figure~\ref{Fig:cycle}. Data points are illustrated by their associated combination of ICD-10CM codes along with the number of patients who experienced them, and the 1-simplexes joining the data points are labeled by the Jaccard distance between the respective sets of patients.
 Therefore, as an example, the label $(R05 \cap R09.3): 33$ means that there are 33 patients who experienced both Cough and Abnormal sputum. 
  The low values of the Jaccard distances imply stronger associations among the respective clusters, which are important to identify the redescriptions.
 In particular, this cycle 
 suggests that among the subjects in $R09.3$--Abnormal sputum--it appears that there is not a particular interaction between subjects in $R05$--Cough--with subjects in $R50.9$--Fever. This opens the question if there is a distinctive signature showing alternative pathways to disease among non-coughers compared to coughers.   

\subsection{Discussion}
To interpret the relationships between the symptoms in Figure~\ref{Fig:cycle} we rely on Jaccard distance. Since the equivalence of sets of subjects matching different patterns produces logical constraints determined by biological processes, multiple pathways connecting phenotypes to disease may yield information about multigenic complex diseases marked by multiple pathways leading to disease.  However, phenotype definitions are prone to misclassification for numbers of reasons.  Therefore, equivalence may be meaningfully characterized by based on the chances that a subject in one or the other of two phenotype clusters is not in both of them, which is the Jaccard distance, described above.
\par
In the case of Figure \ref{Fig:cycle}, there are two paths leading from $R09.3$ (abnormal sputum) to $R05 \cap R09.3 \cap R50.9$, one passing through $R09.3 \cap R50.9$ and the other through $R05 \cap R09.3$, where $R05$ is cough, and $R50.9$ is fever.  In both pathways, the distances between sputum and cough is larger than that between sputum and fever.  So coughing is not as strong an association as fever for abnormal sputum production.  In this case, the relationship between sputum and fever is independent of coughing, since the cycle appears to be a parallelogram.  So a coughing symptom is independent of fever among sputum productive subjects.  This suggests the paths are independent predictors of severe disease.


\section{Conclusions and Future Work} \label{sec:conclusion}

In this study we investigated the application of topological properties in extracting the candidate COVID-19 pathways from the clinical notes. We also proposed a pipeline to pre-process the data, extract the salient concepts, construct a feature space, and visualize the results. We evaluated our pipeline on a set of 1,545 patients and showed that it can extract meaningful associations between symptoms, and reveal intriguing candidate pathways. 

One limitation of our study is the reliance on human validation. As we mentioned in Section \ref{sec:pipeline}, the available text processing tools were not able to effectively parse and extract the relevant medical concepts. To resolve this shortcoming we plan to exploit the available structured data accompanied by the medical records to cluster the patients and automatically filter out the improbable associations.


\begin{small}
\bibliography{ref.bib}

\begin{thebibliography}{19}
\providecommand{\natexlab}[1]{#1}
\providecommand{\url}[1]{\texttt{#1}}
\providecommand{\urlprefix}{URL }
\expandafter\ifx\csname urlstyle\endcsname\relax
  \providecommand{\doi}[1]{doi:\discretionary{}{}{}#1}\else
  \providecommand{\doi}{doi:\discretionary{}{}{}\begingroup
  \urlstyle{rm}\Url}\fi

\bibitem[{Bhatia, Busra~Celikkaya, and Khalilia(2020)}]{Bhatia2020-1}
Bhatia, P.; Busra~Celikkaya, E.; and Khalilia, M. 2020.
\newblock \emph{End-to-End Joint Entity Extraction and Negation Detection for
  Clinical Text}, 139--148.
\newblock Cham: Springer International Publishing.
\newblock ISBN 978-3-030-24409-5.
\newblock \doi{10.1007/978-3-030-24409-5_13}.
\newblock \urlprefix\url{https://doi.org/10.1007/978-3-030-24409-5_13}.

\bibitem[{Dabaghian et~al.(2012)Dabaghian, Mémoli, Frank, and
  Carlsson}]{dabaghian_topological_2012}
Dabaghian, Y.; Mémoli, F.; Frank, L.; and Carlsson, G. 2012.
\newblock A {Topological} {Paradigm} for {Hippocampal} {Spatial} {Map}
  {Formation} {Using} {Persistent} {Homology}.
\newblock \emph{PLOS Computational Biology} 8(8): 1--14.
\newblock \doi{10.1371/journal.pcbi.1002581}.
\newblock \urlprefix\url{https://doi.org/10.1371/journal.pcbi.1002581}.

\bibitem[{Dagliati et~al.(2019)Dagliati, Geifman, Peek, Holmes, Sacchi,
  Sajjadi, and Tucker}]{Dagliati-2019}
Dagliati, A.; Geifman, N.; Peek, N.; Holmes, J.~H.; Sacchi, L.; Sajjadi, S.~E.;
  and Tucker, A. 2019.
\newblock Inferring Temporal Phenotypes with Topological Data Analysis and
  Pseudo Time-Series.
\newblock In Ria{\~{n}}o, D.; Wilk, S.; and ten Teije, A., eds.,
  \emph{Artificial Intelligence in Medicine}, 399--409. Cham: Springer
  International Publishing.
\newblock ISBN 978-3-030-21642-9.

\bibitem[{Demner-Fushman, Chapman, and McDonald(2009)}]{DEMNERFUSHMAN2009760}
Demner-Fushman, D.; Chapman, W.~W.; and McDonald, C.~J. 2009.
\newblock What can natural language processing do for clinical decision
  support?
\newblock \emph{Journal of Biomedical Informatics} 42(5): 760 -- 772.
\newblock ISSN 1532-0464.
\newblock \doi{https://doi.org/10.1016/j.jbi.2009.08.007}.
\newblock
  \urlprefix\url{http://www.sciencedirect.com/science/article/pii/S1532046409001087}.
\newblock Biomedical Natural Language Processing.

\bibitem[{{Edelsbrunner}, {Letscher}, and
  {Zomorodian}(2002)}]{Edelsbrunner2002}
{Edelsbrunner}; {Letscher}; and {Zomorodian}. 2002.
\newblock Topological Persistence and Simplification.
\newblock \emph{Discrete {\&} Computational Geometry} 28(4): 511--533.
\newblock ISSN 1432-0444.
\newblock \doi{10.1007/s00454-002-2885-2}.
\newblock \urlprefix\url{https://doi.org/10.1007/s00454-002-2885-2}.

\bibitem[{Edelsbrunner and Harer(2008)}]{Edelsbrunner_survey}
Edelsbrunner, H.; and Harer, J. 2008.
\newblock Persistent homology---a survey.
\newblock In \emph{Surveys on discrete and computational geometry}, volume 453
  of \emph{Contemp. Math.}, 257--282. Providence, RI: Amer. Math. Soc.

\bibitem[{Jin et~al.(2018)Jin, Bahadori, Colak, Bhatia, Celikkaya, Bhakta,
  Senthivel, Khalilia, Navarro, Zhang, Doman, Ravi, Liger, and
  Kass-Hout}]{ML4H}
Jin, M.; Bahadori, M.~T.; Colak, A.; Bhatia, P.; Celikkaya, B.; Bhakta, R.;
  Senthivel, S.; Khalilia, M.; Navarro, D.; Zhang, B.; Doman, T.; Ravi, A.;
  Liger, M.; and Kass-Hout, T.~A. 2018.
\newblock Improving Hospital Mortality Prediction with Medical Named Entities
  and Multimodal Learning.
\newblock In \emph{Workshop on Machine Learning for Health, NeurIPS}.

\bibitem[{Koleck et~al.(2019)Koleck, Dreisbach, Bourne, and
  Bakken}]{10.1093/jamia/ocy173}
Koleck, T.~A.; Dreisbach, C.; Bourne, P.~E.; and Bakken, S. 2019.
\newblock {Natural language processing of symptoms documented in free-text
  narratives of electronic health records: a systematic review}.
\newblock \emph{Journal of the American Medical Informatics Association} 26(4):
  364--379.
\newblock ISSN 1527-974X.
\newblock \doi{10.1093/jamia/ocy173}.
\newblock \urlprefix\url{https://doi.org/10.1093/jamia/ocy173}.

\bibitem[{Mullins et~al.(2006)Mullins, Siadaty, Lyman, Scully, Garrett, {Greg
  Miller}, Muller, Robson, Apte, Weiss, Rigoutsos, Platt, Cohen, and
  Knaus}]{MULLINS2006}
Mullins, I.~M.; Siadaty, M.~S.; Lyman, J.; Scully, K.; Garrett, C.~T.; {Greg
  Miller}, W.; Muller, R.; Robson, B.; Apte, C.; Weiss, S.; Rigoutsos, I.;
  Platt, D.; Cohen, S.; and Knaus, W.~A. 2006.
\newblock Data mining and clinical data repositories: Insights from a 667,000
  patient data set.
\newblock \emph{Computers in Biology and Medicine} 36(12): 1351 -- 1377.
\newblock ISSN 0010-4825.
\newblock \doi{https://doi.org/10.1016/j.compbiomed.2005.08.003}.
\newblock
  \urlprefix\url{http://www.sciencedirect.com/science/article/pii/S0010482505001046}.

\bibitem[{{Nasrin} et~al.(2019){Nasrin}, {Oballe}, {Boothe}, and
  {Maroulas}}]{8999328}
{Nasrin}, F.; {Oballe}, C.; {Boothe}, D.; and {Maroulas}, V. 2019.
\newblock Bayesian Topological Learning for Brain State Classification.
\newblock In \emph{2019 18th IEEE International Conference On Machine Learning
  And Applications (ICMLA)}, 1247--1252.
\newblock \doi{10.1109/ICMLA.2019.00205}.

\bibitem[{Nicolau, Levine, and Carlsson(2011)}]{nicolau_topology_2011}
Nicolau, M.; Levine, A.~J.; and Carlsson, G. 2011.
\newblock Topology based data analysis identifies a subgroup of breast cancers
  with a unique mutational profile and excellent survival.
\newblock \emph{Proceedings of the National Academy of Sciences} 108(17):
  7265--7270.
\newblock ISSN 0027-8424.
\newblock \doi{10.1073/pnas.1102826108}.
\newblock \urlprefix\url{https://www.pnas.org/content/108/17/7265}.

\bibitem[{Parida and Ramakrishnan(2005)}]{Parida2005RedescriptionMS}
Parida, L.; and Ramakrishnan, N. 2005.
\newblock Redescription Mining: Structure Theory and Algorithms.
\newblock In \emph{AAAI}.

\bibitem[{Platt et~al.(2016)Platt, Basu, Zalloua, and Parida}]{Platt2016}
Platt, D.~E.; Basu, S.; Zalloua, P.~A.; and Parida, L. 2016.
\newblock Characterizing redescriptions using persistent homology to isolate
  genetic pathways contributing to pathogenesis.
\newblock \emph{BMC Systems Biology} 10(1): S10.
\newblock ISSN 1752-0509.
\newblock \doi{10.1186/s12918-015-0251-2}.
\newblock \urlprefix\url{https://doi.org/10.1186/s12918-015-0251-2}.

\bibitem[{Rabad{\'a}n et~al.(2020)Rabad{\'a}n, Mohamedi, Rubin, Chu, Alghalith,
  Elliott, Arn{\'e}s, Cal, Obaya, Levine, and C{\'a}mara}]{Rabadan2020}
Rabad{\'a}n, R.; Mohamedi, Y.; Rubin, U.; Chu, T.; Alghalith, A.~N.; Elliott,
  O.; Arn{\'e}s, L.; Cal, S.; Obaya, {\'A}.~J.; Levine, A.~J.; and C{\'a}mara,
  P.~G. 2020.
\newblock Identification of relevant genetic alterations in cancer using
  topological data analysis.
\newblock \emph{Nature Communications} 11(1): 3808.
\newblock ISSN 2041-1723.
\newblock \doi{10.1038/s41467-020-17659-7}.
\newblock \urlprefix\url{https://doi.org/10.1038/s41467-020-17659-7}.

\bibitem[{Ramakrishnan et~al.(2004)Ramakrishnan, Kumar, Mishra, Potts, and
  Helm}]{10.1145/1014052.1014083}
Ramakrishnan, N.; Kumar, D.; Mishra, B.; Potts, M.; and Helm, R.~F. 2004.
\newblock Turning CARTwheels: An Alternating Algorithm for Mining
  Redescriptions.
\newblock In \emph{Proceedings of the Tenth ACM SIGKDD International Conference
  on Knowledge Discovery and Data Mining}, KDD '04, 266–275. New York, NY,
  USA: Association for Computing Machinery.
\newblock ISBN 1581138881.
\newblock \doi{10.1145/1014052.1014083}.
\newblock \urlprefix\url{https://doi.org/10.1145/1014052.1014083}.

\bibitem[{Wang et~al.(2020)Wang, Hozumi, Yin, and Wei}]{Wang-2020}
Wang, R.; Hozumi, Y.; Yin, C.; and Wei, G.-W. 2020.
\newblock Decoding asymptomatic COVID-19 infection and transmission.

\bibitem[{Wasserman(2018)}]{wasserman-2018}
Wasserman, L. 2018.
\newblock Topological Data Analysis.
\newblock \emph{Annual Review of Statistics and Its Application} 5(1):
  501--532.
\newblock \doi{10.1146/annurev-statistics-031017-100045}.
\newblock
  \urlprefix\url{https://doi.org/10.1146/annurev-statistics-031017-100045}.

\bibitem[{WHO(2020)}]{who}
WHO. 2020.
\newblock \emph{Coronavirus:symptoms}.
\newblock (accessed October 18, 2020),
  \url{https://www.who.int/health-topics/coronavirus}.

\bibitem[{Xu et~al.(2020)Xu, Gutierrez, Mekaru, Sewalk, Goodwin, Loskill, Cohn,
  Hswen, Hill, Cobo, Zarebski, Li, Wu, Hulland, Morgan, Wang, O'Brien,
  Scarpino, Brownstein, Pybus, Pigott, and Kraemer}]{Xu2020}
Xu, B.; Gutierrez, B.; Mekaru, S.; Sewalk, K.; Goodwin, L.; Loskill, A.; Cohn,
  E.~L.; Hswen, Y.; Hill, S.~C.; Cobo, M.~M.; Zarebski, A.~E.; Li, S.; Wu,
  C.-H.; Hulland, E.; Morgan, J.~D.; Wang, L.; O'Brien, K.; Scarpino, S.~V.;
  Brownstein, J.~S.; Pybus, O.~G.; Pigott, D.~M.; and Kraemer, M. U.~G. 2020.
\newblock Epidemiological data from the COVID-19 outbreak, real-time case
  information.
\newblock \emph{Scientific Data} 7(1): 106.
\newblock ISSN 2052-4463.
\newblock \doi{10.1038/s41597-020-0448-0}.
\newblock \urlprefix\url{https://doi.org/10.1038/s41597-020-0448-0}.

\end{thebibliography}
\end{small}

\end{document}